# COMET: Combined Matrix for Elucidating Targets

Haojie Wang#, Zhe Zhang#, Haotian Gao, Xiangying Zhang, Jingyuan Li, Zhihang Chen, Xinchong Chen, Yifei Qi, Yan Li*, and Renxiao Wang*

*Department of Medicinal Chemistry, School of Pharmacy, Fudan University, 826 Zhangheng Road, Shanghai 201203, People's Republic of China*

# These authors had equal contributions to this work.
* To whom all correspondence should be addressed: *li_yan@fudan.edu.cn* (Y. Li); *wangrx@fudan.edu.cn* (R. Wang)

# ABSTRACT

Identifying the interaction targets of bioactive compounds is a foundational element for deciphering their pharmacological effects. Target prediction algorithms equip researchers with an effective tool to rapidly scope and explore potential targets. Here, we introduce the COMET, a multi-technological modular target prediction tool that provides comprehensive predictive insights, including similar active compounds, three-dimensional predicted binding modes, and probability scores, all within an average processing time of less than 10 minutes per task. With meticulously curated data, the COMET database encompasses 990,944 drug-target interaction pairs and 45,035 binding pockets, enabling predictions for 2,685 targets, which span confirmed and exploratory therapeutic targets for human diseases. In comparative testing using datasets from ChEMBL and BindingDB, COMET outperformed five other well-known algorithms, offering nearly an 80% probability of accurately identifying at least one true target within the top 15 predictions for a given compound. COMET also features a user-friendly web server, accessible freely at https://www.pdbbind-plus.org.cn/comet.

# 1. INTRODUCTION

The demand for precise target identification is surging, driven by advances in phenotypic screening[1], the exploration of natural products[2], the scrutiny of compounds' side effects[3], and the reevaluation of marketed drugs for new therapeutic purposes[4]. Identifying all potential interaction targets for a compound enables researchers to better comprehend the pharmacological mechanisms underlying its efficacy and provides direction for subsequent compound optimization. Yet, the inherent limitations of traditional experimental methods are becoming starkly evident. These methods, while varied[5, 6], are often time-consuming, resource-intensive, and constrained by low throughput. In response to these challenges, computational target prediction algorithms emerge as a promising avenue, narrowing down potential targets for active compounds and consequently reducing the experimental workload.

Benefiting from increased data availability, computational capacity, and other key factors, several target prediction algorithms have been developed[7-9]. These algorithms can be broadly categorized into three types based on the information they utilize. The first category comprises ligand-based methods that leverage compound information for predictions, representing the most diverse and widely used approaches in target prediction. Notable examples include the SEA algorithm[10], which performs statistical analysis of the cumulative similarity between the query compound and active molecules associated with each target; SwissTargetPrediction[11], predicting targets based on ligand 2D/3D similarities; SuperPred 3.0[12], which merges machine learning with drug ATC classification; and PPB2, which refines predictions through nearest neighbor searches and Naïve Bayes machine learning techniques[13]. The second category incorporates algorithms that utilize structural information. PharmMapper[14], for instance, reverse docks the query compound into an extensive database of target pharmacophore models, and GalaxySagittarius employs machine learning to match molecules with binding pockets while also offering a combined ligand similarity model for flexible selection by users[15].

The third category features advanced AI-based methods, such as TransformerCPI2.0[16], renowned for their ability to predict targets solely based on protein sequences and ligand SMILES. Another example in this category is CODD-Pred[17], which uses graph neural networks to extract molecular features and integrates machine learning models for both target prediction and affinity prediction.

Despite various advancements in target prediction, certain limitations still hinder their broader adoption. The reference evidence provided by existing methods is often fragmented, compelling researchers to supplement with additional data or conduct further simulations. For instance, ligand-based methods typically only offer data on similar ligands, structure-based methods yield docking results, and AI-based approaches provide merely scoring metrics. Moreover, the absence of performance comparisons among different categories of prediction methods leaves users uncertain about the best choice. To bridge these gaps, we developed COMET, a multi-technological modular target prediction algorithm that combines ligand similarity, protein-ligand docking, AI-based affinity prediction, and machine learning-based ranking algorithms. Leveraging its robust modules, COMET distinguishes itself from other well-known prediction tools by not only providing rich predictive evidence but also achieving an optimal balance between prediction speed and accuracy. Besides this, we have crafted a user-friendly online interface, aiming to offer a comprehensive one-stop prediction service.

## 2. Methods

### 2.1 Compilation of the Data Set

Data from multiple sources were collected and processed to construct COMET. The scope of predictable targets is primarily defined by the Therapeutic Target Database (TTD)[18], focusing on targets that possess both known active compounds and available 3D structures. Human targets not listed in TTD but with the requisite data are also considered, treating them as exploratory within the prediction scope. Protein structures are sourced from the PDBbind[19], PDB[20], and AlphaFold databases[21]. For proteins, aside from those with pockets already identified in PDBbind, binding pockets are predicted using the Cavity tool[22]. Interactions between small molecule ligands and targets are primarily derived from the ChEMBL database[23], with additional data from BindingDB[24] for testing purposes. Details of dataset curation can be found below.

**2.1.1 Protein Structure Dataset**

The protein structure dataset for targets was curated from three sources: PDBbind, PDB, and AlphaFold DB. Priority was given to structures from PDBbind, as these have been rigorously screened and have predefined binding pockets. Next, structures from PDB were used, selected based on the following criteria: non-NMR resolution method, resolution of ⩽2.5 Å, at least 50 amino acid residues, and no covalently bound small molecules. For the remaining targets, predicted structures from AlphaFold DB were used, retaining only those with an average pLDDT score above 70 and cutting disordered loops at the N- and C-termini.

All structures were preprocessed using the Schrödinger Protein Preparation workflow with default parameters. Subsequently, for structures from PDB and AlphaFold DB, the Cavity algorithm was employed to predict binding pockets, retaining up to the top 20 predicted pockets based on Cavity's RankScore and excluding pockets classified as undruggable.

**2.1.2 Ligand Interaction Dataset**

The dataset of small molecule ligand-target interactions was curated from ChEMBL31, with a data acquisition and cleaning protocol encompassing: selection of bioactivity data types including $K_d$, $K_i$, $IC_{50}$, and $EC_{50}$, accepting only activity values with prefixes "=" or "<" and less than 20 μM; processing of ligands involved desalting, retaining the largest fragment in multi-fragment molecules, neutralizing charges, standardizing structures, and deduplicating active molecules per target; integration of data corresponding to human targets from other species (e.g., Mouse, Rat, Bovine) into the dataset; and segmentation of the data into three subsets following the Similarity Ensemble Approach (SEA) algorithm[25], based on the number of active molecules per target: fewer than 5, between 5 and 300, or more than 300.

Moreover, newly identified compound-target interactions from ChEMBL32 were organized for parameter tuning in the ligand module. Multi-target compound data were also curated from ChEMBL32 for model training and testing in the ranking module, while ChEMBL33 and BindingDB data were used for overall performance evaluation of the COMET algorithm. For the multi-target compound, additional criteria were applied: a maximum of 100 heavy atoms; exclusion of elements other than H, C, N, O, P, S, F, Cl, Br, I, Mg, Ca, Zn, Fe, Mn; at least two validated targets per compound; and retaining only one representative compound per scaffold, defined using the Murcko framework[26], for compounds with the same targets to reduce redundancy. For the BindingDB data, selections were restricted to entries labeled "Curated from the literature by BindingDB" under the "Curation/DataSource" tag.

**2.2 COMET Framework**

We constructed a multi-technology modular algorithm architecture that includes the following components: a ligand-based initial target screening module, which combines cumulative similarity analysis, maximum Tanimoto similarity, and clustering-based target association for initial target screening; an AI-based affinity prediction module using our in-house developed tool, PLANET[27], to efficiently predict affinities between query compounds and all protein pockets of screened targets, also assessing the affinity of positive and random compounds as references for subsequent ranking; a machine learning-based target ranking module that integrates scores from both the ligand and affinity modules to provide confidence scores for initial target predictions, ranking the top 100 targets; and a docking module that utilizes the well-established Autodock Vina GPU version for suggesting possible protein-ligand binding modes.

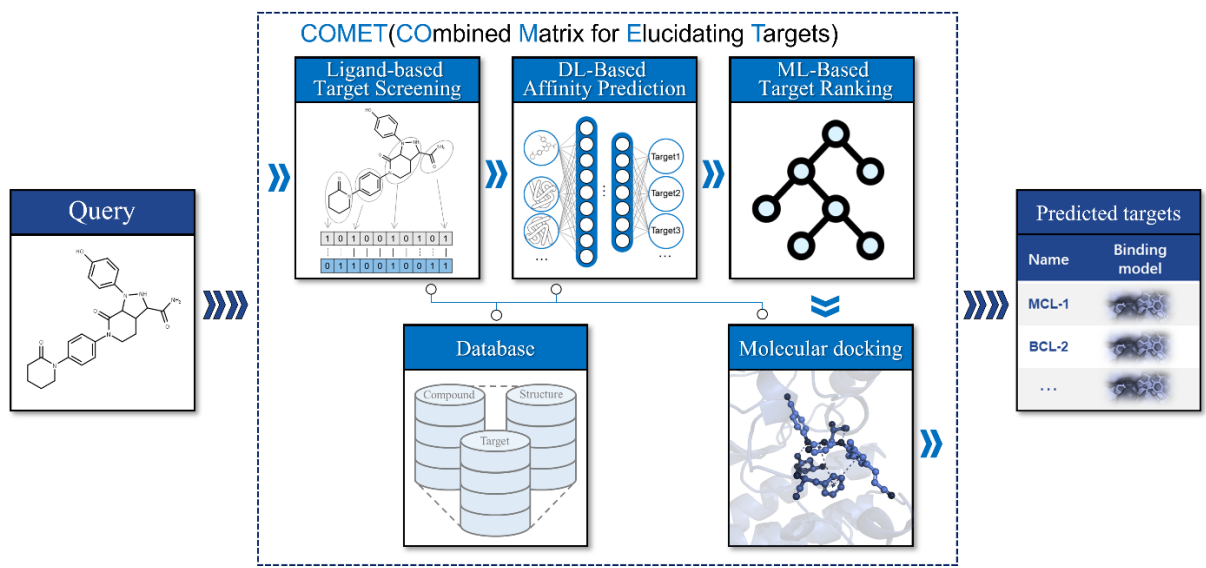

**Figure 1.** Framework of the COMET Algorithm. For the input query compound structure, the ligand-based module performs initial target screening. The AI-based module then predicts the affinities between the compound and the initially screened targets. And the machine learning-based module uses scores from the first two modules to rank the predicted targets. Subsequently, the final module supplements docking information for each predicted target.

The overall workflow of the COMET algorithm is illustrated in Figure 1. Starting with small molecule structures as inputs, it produces outputs that include predicted probabilities, structures of similar bioactive compounds, and observable predicted binding modes.

## 2.2.1 Ligand Module

Our ligand module employs a combination of cumulative similarity analysis, clustering-based target association, and maximum similarity analysis for initial target screening. The essence of the first approach is constructing a statistical model to assess if the cumulative similarity between two sets of compounds is statistically significant compared to that of a similarly sized but randomly sampled compound set. This model is based on Keiser et al.'s SEA algorithm and its variants[10, 25, 28, 29]. To accommodate the statistical model, the ligand interaction dataset is divided into three subsets based on the number of known active molecules for each target, with separate statistical models constructed for each subset accordingly.

*Subset 1: Targets with 5 to 300 Known Active Molecules.* To simulate the number of compound pairs analyzed when calculating cumulative similarity across different-sized compound sets, 1,000 random integers between $10^2$ and $300^2$ are sampled. For each integer $S$, compounds are randomly selected from the subset to form two sets, A and B, with $i$ and $j$ compounds respectively, where $i \times j = S$ and $i, j \in (10, 300)$. Then, the Raw Scores are calculated using Equations 1. In the formula, the similarity is calculated using the Tanimoto coefficient ($Tc$) with ECFP4 fingerprints, and $Ts$ represents the similarity threshold. This step is repeated 30 times for each $S$.

$$Rawscore(A, B) = \sum_i \sum_j simi(A_i, B_j)$$

$$simi(A_i, B_j) = \begin{cases} Tc(A_i, B_j) & \text{if } Tc(A_i, B_j) \geq Ts \\ 0 & \text{if } Tc(A_i, B_j) < Ts \end{cases} \quad \text{(Equation 1)}$$

As $S$ increases, Raw Scores are expected to increase, with mean and variance functions fitted using Equation 2. Further statistic values are obtained using Equations 3-4, converting raw scores to Z-scores (z) and then to P-values. P-values < 0.05 indicate significance.

$$F_{\text{mean|std}}(S) = uS^r + c \quad \text{(Equation 2)}$$

$$Z\text{-}score = \frac{Rawscore - F_{\text{mean}}(S)}{F_{\text{std}}(S)} \quad \text{(Equation 3)}$$

$$P\text{-}value(z) = \begin{cases} 1 - e^{X(z)} & \text{if } z \leq 28 \\ -X(z) - \frac{X(z)^2}{2} - \frac{X(z)^3}{6} & \text{if } z > 28 \end{cases}$$

$$X(z) = -e^{-\frac{z\pi}{\sqrt{6} - 0.577215665}} \quad \text{(Equation 4)}$$

In this model-building process, $Ts$ is sampled at intervals of 0.01 from 0 to 1, generating 100 sets of fitting parameters. We employ an empirical method to identify the optimal $Ts$ value, using newly added active compound data from corresponding targets in ChEMBL32 as a training set, and select the $Ts$ value that yields the highest recall performance. The chosen $Ts$ values and other fitting parameters are provided in Supplementary Information Table S1.

*Subset 2: Targets with More than 300 Known Active Molecules.* Before parameter fitting, compounds are clustered by Murcko scaffold[26], retaining one representative molecule per scaffold for each target. Parameters are then fitted using sampled integer $S$ values ranging from $100^2$ to $2000^2$, following a similar process as in Subset 1.

*Subset 3: Targets with Fewer than 5 Known Active Molecules.* The parameter fitting method is slightly adjusted: Compounds are randomly selected to form sets A and B, iterating over compound numbers $i$ and $j$ from 1 to 50. For each $i \times j$ combination, the sampling and raw score calculation process is repeated 100 times, with mean and standard deviation fitted using the modified Equations 5 and 6. Z-score and P-value calculations remain the same.

$$F_{mean}(S) = uS \qquad \text{(Equation 5)}$$

$$F_{std}(S) = qS^r \qquad \text{(Equation 6)}$$

Across all three subsets, similarity is calculated via the Rdkit package[30]. The mean and standard deviation are fitted using the least squares method from the SciPy package[31].

For the second clustering-based target association method, targets are pre-clustered using a fine-tuned cumulative similarity analysis model. Upon identifying a target by the cumulative similarity analysis, its associated targets are also considered. Compared to the first method, the clustering statistical model has two adjustments: first, it follows Keiser et al.'s original SEA method[10] by fitting Z-scores to an extreme value distribution (EVD) using Scipy's stats.gumbel_r. The fit is assessed using a chi-square test to determine the appropriate $Ts$ value, which enhances the statistical reliability of the clustering model. Second, calculating cumulative similarity between targets and sorts associated targets based on their E-values. To prevent excessive associations, each target is linked to up to three others, with required E-values of less than $10^{-10}$, $10^{-50}$, and $10^{-100}$, respectively.

$$E\text{-}value(z) = P\text{-}value(z) * N_{db} \qquad (\text{Equation 7})$$

where $N_{db}$ is the number of set comparisons made in the subset.

For the third maximum similarity method, similarities between the query compound and all known active molecules of the targets are calculated. Each target retains the highest similarity value, and targets with similarity values greater than 0.4 are selected, an empirical threshold based on prior studies[25].

**2.2.2 Affinity Prediction and Molecular Docking Module**

PLANET is a deep learning protein-ligand affinity prediction algorithm developed by our research group, based on Graph Neural Networks (GNN). This method requires only the three-dimensional structure of the binding pocket and the SMILES of the compound to predict affinity. In this work, we use PLANET to predict the affinity between query compounds and the binding pockets of targets initially screened by the ligand module. Typically, a target may have multiple crystals in the database, corresponding to multiple experimentally validated or predicted binding pockets. For each target, the binding pocket with the highest predicted affinity is retained, along with the predicted affinity values.

In addition to the query compounds, we assess the affinity of known positive compounds and background compounds against the target's binding pocket. The background compounds come from the National Cancer Institute (NCI) Diversity Set VI, which includes 1,584 structurally diverse compounds and are treated as negative background molecules. The average predicted affinity values for both the positive and background compounds are used as reference points.

Subsequently, molecular docking is performed with Vina-GPU 2.0[32], a GPU-accelerated version of AutoDock Vina, which facilitates the docking of query compounds within the pockets. This provides a visual representation of the predicted binding modes and molecular docking scores for user verification.

**2.2.3 Ranking Module**

Predictive scores from both the ligand module and the affinity module are integrated using a random forest classification model, which outputs the probability of each predicted target being a true positive. The predicted targets are then ranked based on these classification probabilities. To train and test this model, we selected 1,206 multi-target compounds from ChEMBL32 and processed them through the ligand and affinity prediction modules to collect predicted targets and their corresponding scores. These data were split into a training set (80%) and a test set (20%) based on compounds.

During the model development phase, we evaluated three machine learning methods: Random Forest (RF), Gradient Boosting Decision Trees (GBDT), and XGBoost. Based on three key metrics—ROC-AUC, recall for the top 100 predictions, and performance in predicting the top 15 targets—the RF model was slightly superior compared to the other two models (a comparative analysis of these models' performances is provided in Supplementary Information Table S3). Consequently, the RF model was chosen for the ranking module. A refined RF model was then trained using the complete dataset of multi-target compounds to optimize its performance.

**2.3 Web Server Implementation**

The COMET server is publicly accessible through web browsers and supports major browsers such as Edge and Chrome. It is built using the Vue[33] framework and includes both frontend and computation nodes, deployed on Tencent Cloud. The task system is implemented using MySQL[34] in conjunction with Redis Queue, small molecule display is managed with the Ketcher[35] plugin, and docking result visualization is handled by the Molstar[36] plugin.

# 3. Results and Discussion

**3.1 COMET Dataset Profiling**

COMET focuses specifically on targets with therapeutic potential for human diseases. The scope of predictable targets has been meticulously compiled, primarily based on the Therapeutic Target Database (TTD), encompassing 2685 targets, 2606 of which are human-derived and 79 from non-human pathogenic organisms. While accuracy remains a priority, users of target prediction tools generally seek support for a broader range of predictable targets. As detailed in Table 1, COMET's target count is highly competitive relative to other platforms. Although platforms like SEA and SwissTargetPrediction feature larger numbers, they unnecessarily differentiate between homologous targets. For example, SwissTargetPrediction supports 3068 targets, but only 2092 are human-derived, with the remaining 976 sourced from rats and mice. At COMET, we recognize the relevance of data from animal models for human disease research and thus integrate homologous targets and their associated bioactive compound data.

**Table 1.** Comparative Analysis of Web-Based Tools for Computational Target Prediction

| Web server | Data sets | | Output Reference Evidence | |
|---|---|---|---|---|
| | Targets | Interactions | Similar bioactive compound | Predictive binding mode |
| COMET | 2685 | 990944 | ✓ | ✓ |
| SwissTargetPrediction | 3068 | 580496 | ✓ | × |
| SEA | 4685 | 796069 | ✓ | × |
| PPB2 | 1720 | 555346 | ✓ | × |
| SuperPred 3.0 | 2353 | 500979 | ✓ | × |
| CODD-Pred | 640 | 874371 | ✓ | × |
| PharmMapper | 1627 | 52431[a] | × | ✓ |
| GalaxySagittarius | 6037 | 389237 | × | ✓ |

Furthermore, unlike GalaxySagittarius, which supports a wide variety of targets including those with only structural data, COMET imposes stringent requirements for known active molecules and reliable protein structures. This results in a repository that includes 990,944 protein-ligand interactions and 45,035 experimentally solved or predicted small molecule binding pockets. Consequently, COMET not only provides richer predictive evidence but also stands out as the only platform offering detailed data on similar active compounds and predicted binding modes. This comprehensive information is crucial, reducing the need for supplementary external data gathering and enabling users to quickly assess each prediction's reliability using their professional expertise.

## 3.2 Performance on Two Test Sets

To rigorously evaluate COMET's target prediction capabilities, we randomly selected 500 compounds from the latest ChEMBL33 database that were not previously included in COMET's dataset. These compounds were required to meet the following criteria: having at least two experimentally verified targets with activities ($K_i$, $K_d$, $IC_{50}$, or $EC_{50}$) under 20 µM, containing no more than 100 heavy atoms, and excluding atom types unsupported by Autodock Vina. In comparing COMET against five popular target prediction algorithms, performance benchmarks akin to those employed by SwissTargetPrediction and GalaxySagittarius were utilized[11, 15]. As illustrated in Figures 2A and 2B, COMET demonstrated superior performance, recalling 1,138 out of 1,456 protein-ligand interactions from the top 100 predicted targets, achieving a recall rate of 72.18%. For each tested compound, COMET had a 77.8% probability of identifying at least one true target within the top 15 predictions.

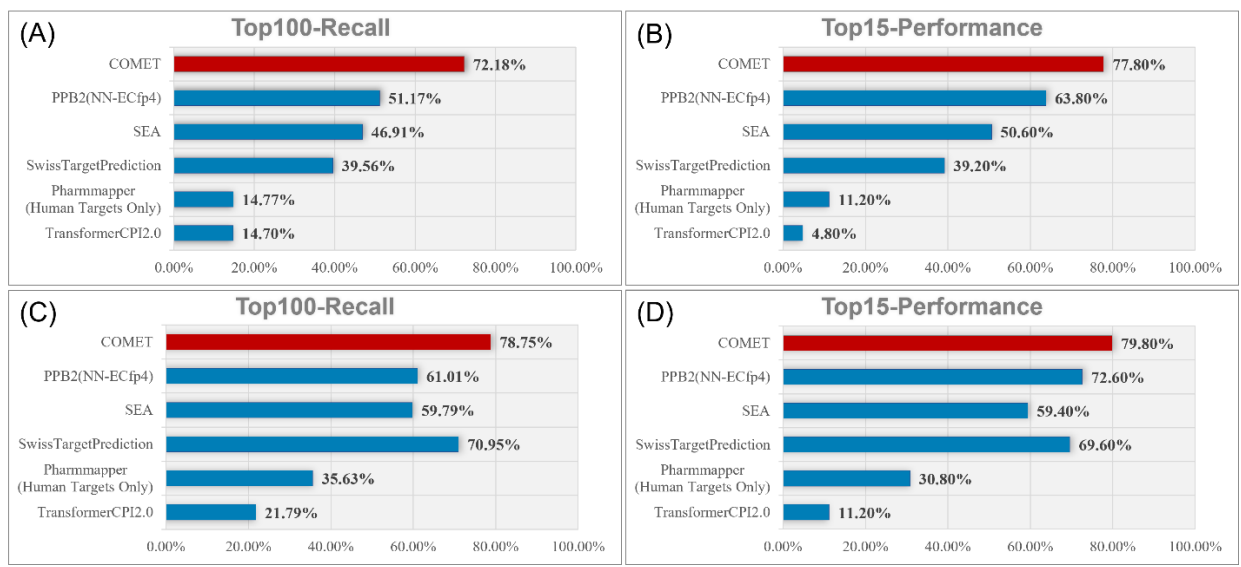

**Figure 2.** Comparison of Target Prediction Performance between COMET and Five Other Algorithms. (A) Top 100 Recall and (B) Top 15 Performance for the test set from ChEMBL33. (C) Top 100 Recall and (D) Top 15 Performance for the test set from BindingDB.

Further robustness testing was conducted with an extra set of 500 compounds from the external BindingDB database, meeting the same criteria. As shown in Figures 2C and 2D, all algorithms showed improvement, attributable to the relatively lower prediction challenges of this task. Notably, SwissTargetPrediction exhibited the greatest enhancement, with a 30% increase in Top 15 Performance, benefiting from an increased number of similar active compounds used as references in the prediction process. In contrast, COMET's improvement was more modest, with only a 2% increase in Top 15 Performance. Detailed examination revealed that COMET accurately predicted true targets for 31 compounds among the set of Top 15 prediction failures, yet the ranking module faced difficulties differentiating them due to several potential targets receiving closely matched or better scores from multiple modules. Despite these challenges, COMET still outperformed other models on both key testing metrics. Moreover, it was observed that AI-based models like TransformerCPI2.0 and structure-based models PharmMapper demonstrated relatively poor performances, highlighting the current limitations of relying solely on AI scoring or docking results to precisely identify true targets from a vast pool of potential candidates.

### 3.3 Ablation Experiments and Feature Importance Analysis

We developed five variants of COMET to explore the impact of each component on its performance. As illustrated in Figures 3A and 3B, the ligand module proved essential for COMET's overall functionality. Omitting this module and relying solely on the affinity and ranking modules for prediction led to a slight decline in Top 100 Recall but a significant (≈35%) decrease in Top 15 Performance.

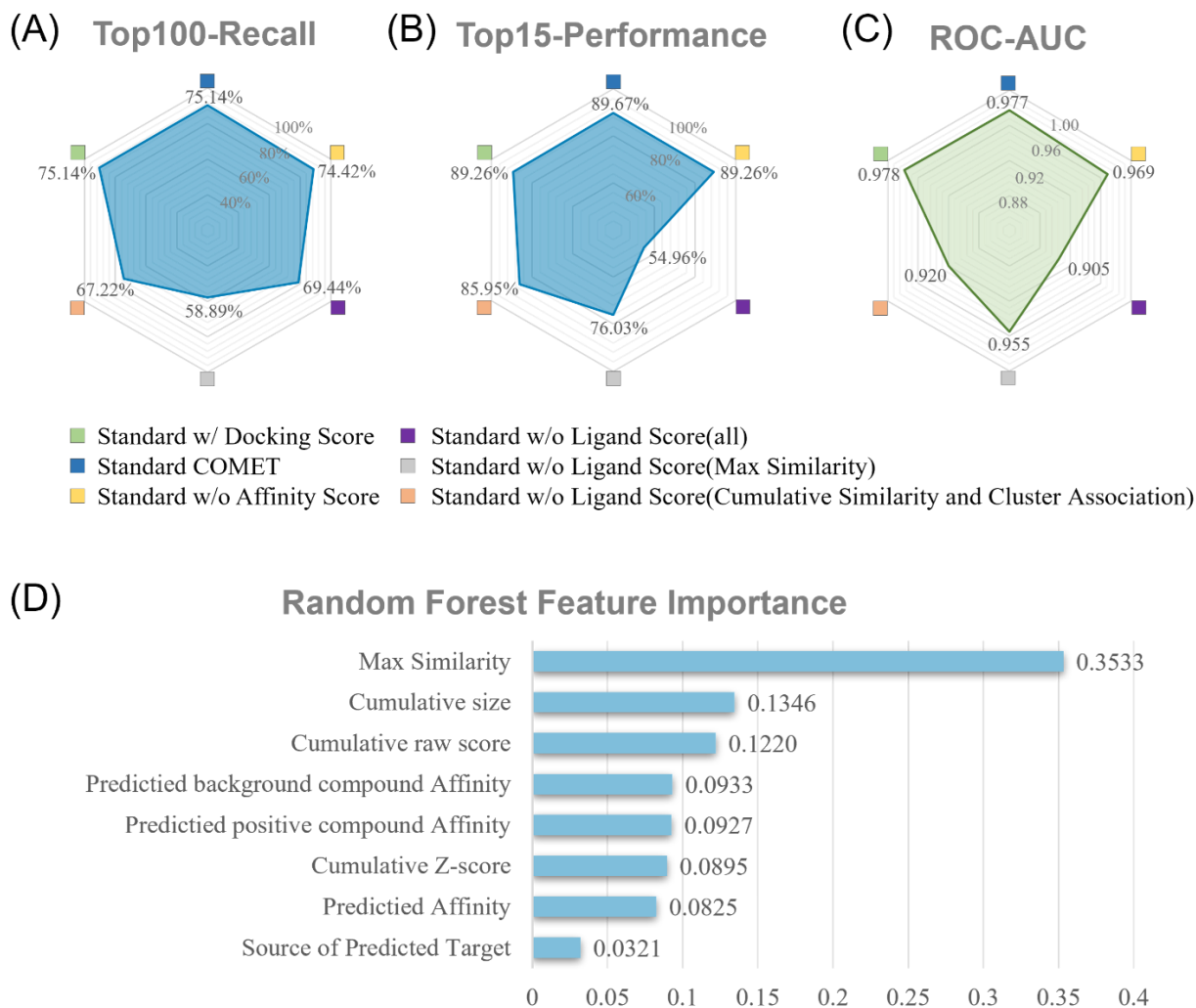

**Figure 2.** Ablation Study and Feature Importance Analysis Results. (A) Top 100 Recall and (B) Top 15 Performance results for COMET and its five variants, based on independent predictions and rankings

for 242 test compounds by each variant. (C) ROC-AUC results of COMET and its five variants, based on predictions for 242 ChEMBL32 test compounds using the standard COMET model, with different module scores masked during ranking. (D) Feature importance analysis results for the ranking module.

An analysis of the three initial screening algorithms within the ligand module revealed that cumulative similarity and target association information had a substantial impact on the ranking capability, as detailed by the ROC-AUC (Figures 3C). On the other hand, the absence of the Max Similarity feature resulted in significant reductions in both Top 100 and Top 15 metrics, highlighting its vital role in initial screening and the subsequent ranking of targets. Feature importance analysis (Figures 3D) also supported this, with Max Similarity showing the highest importance ratio of 0.35.

Affinity scoring modestly improved the ROC-AUC metric from 0.969 to 0.977, and the combined importance of the three affinity prediction metrics totaled approximately 0.27, both affirming their substantive contributions to the model's accuracy. However, docking scores offered minimal enhancement in practical performance metrics like Top 15 Performance and Top 100 Recall, leading to their exclusion from the ranking module considerations.

**3.4 Web Interface Usage**

COMET provides a user-friendly interface for researchers. For a demonstration, we utilized the FDA-approved drug Vilazodone, primarily known as a selective serotonin reuptake inhibitor (SSRI) and a partial agonist at the 5-HT1A receptor. Intriguingly, it also exhibited activity against additional targets such as the Dopamine transporter, Norepinephrine transporter, and hERG, with $IC_{50}$ values of 295.0 nM, 158.1 nM, and 5.55 μM, respectively. The compound's SMILES was entered into the submission interface (Figure 4A), which accommodates file uploads and drawing board editing. After entering a job name, clicking the prediction button initiates the job submission process.

Subsequently, users can monitor all their job statuses on the Job Status page without the need for common email notifications (Figure 4B). This page displays total queue numbers, estimated waiting times, and the status of submitted tasks, with each task requiring about 10 minutes of computation. Upon completion, the "View Results" option becomes available. The results page, as shown in Figure 4C, presents probability values, molecular docking results, and the structures of similar active compounds for each prediction. To enable detailed examination, docking results are showcased in four distinct display styles, and up to 20 reference active compounds per target can be viewed by sliding left and right. Probability levels range from 0 to 100%, with values near 100% typically indicating experimentally verified targets. For Vilazodone, the top 12 targets were all previously confirmed and are recorded in the COMET database, while the top 13-15 accurately predicted the dopamine transporter, norepinephrine transporter, and hERG, underscoring COMET's predictive accuracy. Additionally, the results page provides organized links to databases for target proteins, genes, active compounds, structures, and related diseases, facilitating rapid access to detailed information about the targets of interest.

**Figure 4.** Prediction Demonstration of Vilazodone on the COMET Website. (A) Uploading the structure of Vilazodone on the input page. (B) Task status overview page. (C) Detailed information on the prediction results page.

## 4. Conclusion

We developed COMET, a target prediction algorithm that is more aligned with user needs, offering rich predictive information, ease of use, and high accuracy. The algorithm focuses on predicting targets with therapeutic potential for human diseases and effectively combines several powerful technologies. It begins with classical ligand similarity methods for initial target screening, then incorporates AI-based affinity prediction through PLANET and molecular docking using Autodock Vina to consider structural information. The AI component not only reduces the computational burden of docking but also provides valuable affinity predictions. Finally, a machine learning ranking algorithm integrates various features to optimally rank the predicted targets. Compared to other well-known methods, COMET demonstrates superior predictive performance. Nonetheless, COMET faces certain limitations, as it currently depends on ligand and structural data, which constrains its predictive scope. In future updates, we will broaden the ligand and structural databases to expand the range of predictable targets, while actively exploring new approaches to incorporate diverse data types, aiming to further elevate the algorithm's predictive capabilities.

## Supporting Information

Detailed parameters of the Rawscore and Z-score formulas across different subsets for target clustering and cumulative similarity analysis (Table S1); parameter search ranges and optimization results for Random Forest, GBDT, and XGBoost algorithms (Table S2); ranking performance of Random Forest, XGBoost, and GBDT on test sets from ChEMBL32 (Table S3).

## Data and Software Availability

The range of predictable targets supported by COMET and the algorithm's testing datasets will be available on GitHub soon.

# TABLE OF CONTENTS GRAPHICS

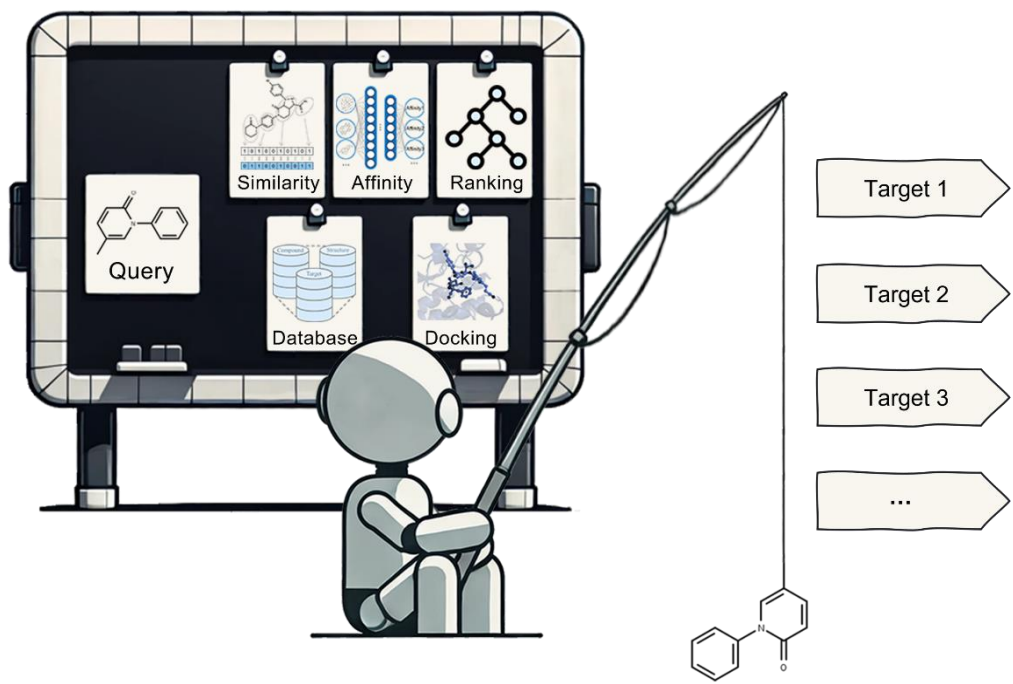

## *Supporting Information*

**Table S1.** Rawscore and Z-score Formulas Across Different Subsets for Target Clustering and Cumulative Similarity Analysis

| Subset | For Target Clustering | For cumulative similarity analysis |
|---|---|---|
| | $Rawscore(A, B) = \sum_i \sum_j simi(A_i, B_j)$ | |
| Subset 1 | $simi(A_i, B_j) = \begin{cases} Tc(A_i, B_j) & \text{if } Tc(A_i, B_j) \geq 0.19 \\ 0 & \text{if } Tc(A_i, B_j) < 0.19 \end{cases}$ | $simi(A_i, B_j) = \begin{cases} Tc(A_i, B_j) & \text{if } Tc(A_i, B_j) \geq 0.25 \\ 0 & \text{if } Tc(A_i, B_j) < 0.25 \end{cases}$ |
| Subset 2 | $simi(A_i, B_j) = \begin{cases} Tc(A_i, B_j) & \text{if } Tc(A_i, B_j) \geq 0.19 \\ 0 & \text{if } Tc(A_i, B_j) < 0.19 \end{cases}$ | $simi(A_i, B_j) = \begin{cases} Tc(A_i, B_j) & \text{if } Tc(A_i, B_j) \geq 0.18 \\ 0 & \text{if } Tc(A_i, B_j) < 0.18 \end{cases}$ |
| Subset 3 | $simi(A_i, B_j) = \begin{cases} Tc(A_i, B_j) & \text{if } Tc(A_i, B_j) \geq 0.5 \\ 0 & \text{if } Tc(A_i, B_j) < 0.5 \end{cases}$ | $simi(A_i, B_j) = \begin{cases} Tc(A_i, B_j) & \text{if } Tc(A_i, B_j) \geq 0.24 \\ 0 & \text{if } Tc(A_i, B_j) < 0.24 \end{cases}$ |
| | $Z\text{-}score = \dfrac{Rawscore - F_{mean}(S)}{F_{std}(S)}$ | |
| Subset 1 | $F_{mean}(S) = 1.23 \times 10^{-2} \times S^{0.999} - 0.187$<br>$F_{std}(S) = 1.63 \times 10^{-2} \times S^{0.728} + 1.36$ | $F_{mean}(S) = 3.1 \times 10^{-3} \times S^{0.983} - 0.818$<br>$F_{std}(S) = 1.45 \times 10^{-2} \times S^{0.652} + 0.273$ |
| Subset 2 | $F_{mean}(S) = 1.11 \times 10^{-2} \times S$<br>$F_{std}(S) = 3.78 \times 10^{-2} \times S^{0.617}$ | $F_{mean}(S) = 1.5 \times 10^{-2} \times S$<br>$F_{std}(S) = 4.12 \times 10^{-2} \times S^{0.632}$ |
| Subset 3 | $F_{mean}(S) = 9.54 \times 10^{-5} \times S^{0.999} - 0.108$<br>$F_{std}(S) = 1.47 \times 10^{-3} \times S^{0.64} + 0.108$ | $F_{mean}(S) = 2.92 \times 10^{-3} \times S^{0.999} - 1.61$<br>$F_{std}(S) = 5.6 \times 10^{-3} \times S^{0.725} + 4.87$ |

**Table S2.** Parameter Search Ranges and Optimization Results for Random Forest, GBDT, and XGBoost Algorithms

| Methods | Parameter | Search Range | Optimal Value |
|---|---|---|---|
| **Random Forest** | n_estimators | $[100, 1000] \cap \mathbb{Z}$ | 610 |
| | max_depth | $[3, 30] \cap \mathbb{Z}$ | 26 |
| | min_samples_split | $[2, 20] \cap \mathbb{Z}$ | 7 |
| | min_samples_leaf | $[1, 20] \cap \mathbb{Z}$ | 2 |
| | bootstrap | [True, False] | False |
| | max_features | [sqrt, log2, None] | sqrt |
| **GBDT** | n_estimators | $[100, 1000] \cap \mathbb{Z}$ | 445 |
| | learning_rate | [0.01, 0.30] | 0.061 |
| | max_depth | $[3, 10] \cap \mathbb{Z}$ | 7 |
| | min_samples_split | $[2, 20] \cap \mathbb{Z}$ | 13 |
| | min_samples_leaf | $[1, 20] \cap \mathbb{Z}$ | 9 |
| | Subsample | [0.5, 1] | 0.947 |
| | max_features | [sqrt, log2, None] | log2 |
| | min_impurity_decrease | [0, 0.1] | 0.012 |
| **XGBoost** | n_estimators | $[100, 1000] \cap \mathbb{Z}$ | 808 |
| | learning_rate | [0.01, 0.30] | 0.026 |
| | max_depth | $[3, 10] \cap \mathbb{Z}$ | 8 |
| | min_child_weight | $[1, 10] \cap \mathbb{Z}$ | 1 |
| | gamma | [0, 0.5] | 0.137 |
| | subsample | [0.5, 1] | 0.778 |
| | colsample_bytree | [0.5, 1] | 0.707 |
| | reg_alpha | [0, 1] | 1.000 |
| | reg_lambda | [0, 1] | 0.997 |

**Table S3**. Ranking Performance of Random Forest, XGBoost, and GBDT on Test Sets from ChEMBL32

|  | ROC-AUC | Top 100 Recall[a] | Top 15 Performance[b] |
|---|---|---|---|
| **Random Forest** | 0.977 | 75.14% | **89.67%** |
| **XGBoost** | **0.979** | **75.42%** | 88.84% |
| **GBDT** | 0.974 | 75.14% | 88.43% |

[a]The recall rate of the top 100 predicted targets for true drug-target pairs.

[b]The probability that at least one true target is hit among the top 15 predicted targets for the compound.